\documentclass[letterpaper]{article} 
\usepackage{aaai2026}  
\usepackage{times}  
\usepackage{helvet}  
\usepackage{courier}  
\usepackage[hyphens]{url}  
\usepackage{graphicx} 
\usepackage{natbib}  
\usepackage{caption} 
\usepackage{algorithm}
\usepackage{algorithmic}
\usepackage{algorithm}
\usepackage{algorithmic}
\usepackage{booktabs}
\usepackage{multirow}
\usepackage{graphicx}
\usepackage{amsmath}
\usepackage{multirow}
\usepackage{graphicx}
\usepackage{stix,bbding,pifont,utfsym,fontawesome}
\usepackage{xcolor}

\usepackage{newfloat}
\usepackage{listings}
\DeclareCaptionStyle{ruled}{labelfont=normalfont,labelsep=colon,strut=off} 
\floatstyle{ruled}
\newfloat{listing}{tb}{lst}{}
\floatname{listing}{Listing}
\title{Crossing Borders: A Multimodal Challenge for Indian Poetry Translation and Image Generation}
\author{
    Sofia Jamil\textsuperscript{\rm 1},
    Kotla Sai Charan\textsuperscript{\rm 1},
    Sriparna Saha\textsuperscript{\rm 1},
    Koustava Goswami\textsuperscript{\rm 2},
    Joseph K J\textsuperscript{\rm 2}
}
\affiliations{
    \textsuperscript{\rm 1}Department of Computer Science and Engineering, Indian Institute of Technology Patna, India\\

    \textsuperscript{\rm 2}Adobe Research\\
}

\usepackage{bibentry}

\begin{document}

\maketitle

\begin{abstract}
Indian poetry, known for its linguistic complexity and deep cultural resonance, has a rich and varied heritage spanning thousands of years. However, its layered meanings, cultural allusions, and sophisticated grammatical constructions often pose challenges for comprehension, especially for non-native speakers or readers unfamiliar with its context and language. Despite its cultural significance, existing works on poetry have largely overlooked Indian language poems. In this paper, we propose the \textbf{Translation and Image Generation (TAI)} framework, leveraging Large Language Models (LLMs) and Latent Diffusion Models through appropriate prompt tuning. Our framework supports the United Nations Sustainable Development Goals of Quality Education (SDG 4) and Reduced Inequalities (SDG 10) by enhancing the accessibility of culturally rich Indian-language poetry to a global audience. It includes (1) a translation module that uses an Odds Ratio Preference Alignment Algorithm to accurately translate morphologically rich poetry into English; (2) an image generation module that employs a semantic graph to capture tokens, dependencies, and semantic relationships between metaphors and their meanings, to create visually meaningful representations of Indian poems. Our comprehensive experimental evaluation, including both human and quantitative assessments, demonstrates the superiority of \textit{TAI} Diffusion in poem image generation tasks, outperforming strong baselines. To further address the scarcity of resources for Indian-language poetry, we introduce the \textbf{Morphologically Rich Indian Language Poems \textit{MorphoVerse} Dataset}, comprising 1,570 poems across 21 low-resource Indian languages. By addressing the gap in poetry translation and visual comprehension, this work aims to broaden accessibility and enrich the reader’s experience. 
\end{abstract}

\begin{links}
    \link{Code}{https://github.com/SofeeyaJ/Crossing-Borders---AAAI26}
\end{links}

\section{Introduction}

Indian poetry and culture is a rich and diverse form of literature that includes several languages and countless dialects, each with its own distinct style, rhythm, and cultural significance. From \textit{Kalidasa's} classical \textit{Sanskrit} verses to the lyrical compositions of \textit{Tamil Sangam} poetry, \textit{Urdu} ghazals to \textit{Bengali} Poems on Social Causes, Indian poetry reflects a diverse range of emotions, philosophies, and artistic expressions.
The morphological richness of Indian languages, with their complex word structures and intricate grammatical rules, makes poetry deeply expressive but challenging to comprehend across languages \cite{maji2025sanskriticomprehensivebenchmarkevaluating,maji-etal-2025-drishtikon}. They sometimes create a barrier to perception, making it difficult for modern readers to fully appreciate and recognize the works of great poets, often leading to unintended neglect of these literary treasures. This necessitates the need for a generalized framework capable of interpreting poetry in a way that is accessible to a general reader unfamiliar with Indian languages. Motivated by this, and in alignment with the United Nations Sustainable Development Goals of Quality Education (SDG 4) and Reduced Inequalities (SDG 10), we propose \textbf{Translation and Image Generation (TAI)}, a two-step framework that translates morphologically rich, low-resource Indian poetry and generates images that visually convey their meanings.

\begin{figure}[!ht]
\centerline{\includegraphics[width=\columnwidth]{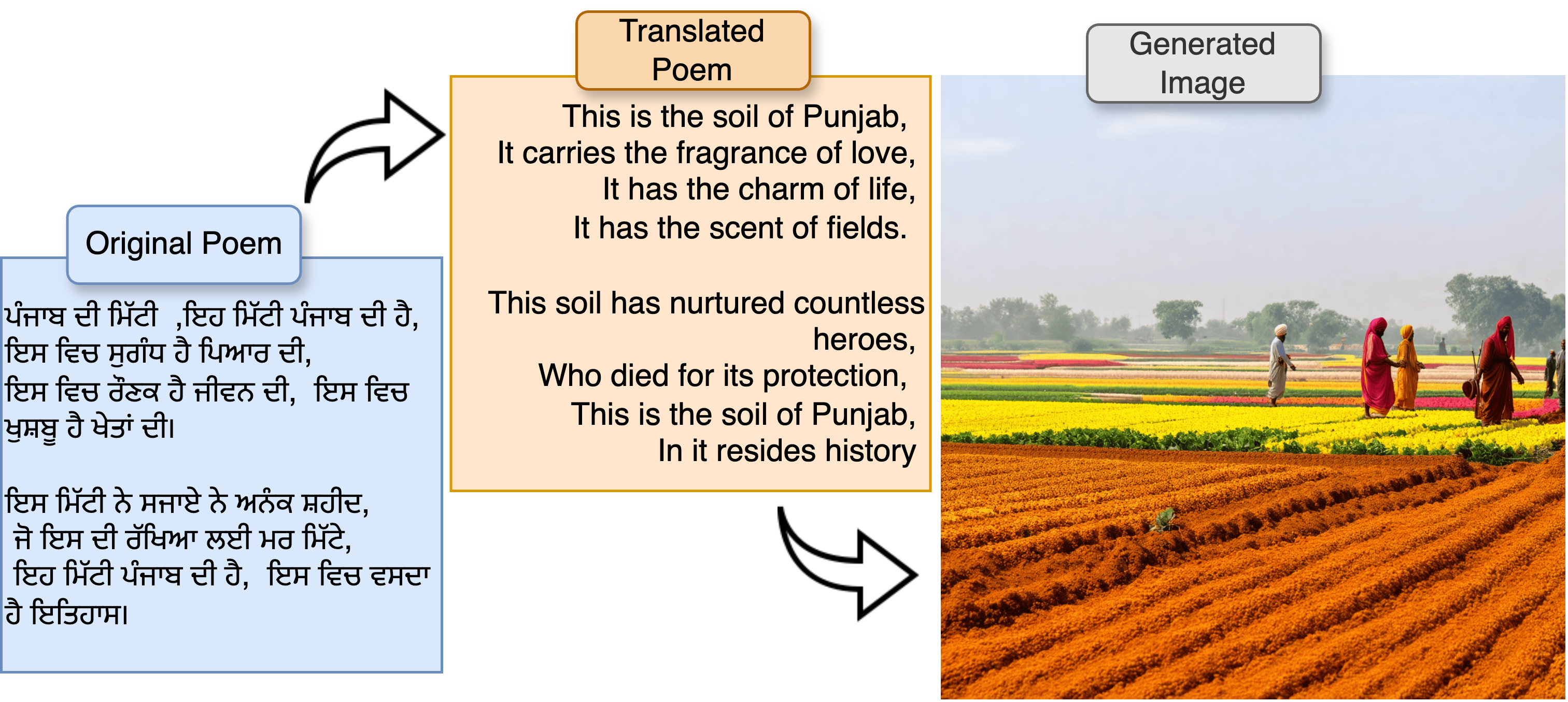}}
\caption{An example of our \textbf{Translation and Image Generation \emph{TAI}} framework, which first translates a Punjabi poem (a low-resource Indian language) into English and then generates a visual representation, capturing the poem's key visual elements (A field with farmers dressed in traditional Punjabi attire).}
\label{motivation_image}
\end{figure}

To date, previous research has explored machine translation for poetry in an excellent way, but most studies have focused on structured poetic forms with clear format or rhyme constraints \cite{ref2,ref3,ref5}, which differ significantly from the morphologically rich and structurally diverse nature of Indic poetry. A comparative study on machine vs. human translation of Arabic poetry into English (\cite{ref6}) concluded that machine translation struggles to capture the socio-cultural context and nuanced poetic elements, making it unsuitable for literary translation.

Our research question is: \textit{How can we interpret Indian poetry without losing its poetic meanings embedded in the verses?} With the introduction of transformer-based architectures and Large Language Models (LLMs), they have exhibited remarkable capabilities in Natural Language Processing (NLP) tasks involving translations and beyond \cite{ref7}. However, while LLMs have demonstrated improved performance in English translation, their effectiveness in Indic languages remains limited \cite{ref8}, and their performance in poetry translation is often inadequate. Consequently, in our research, we incorporated an alignment algorithm to guide LLMs in generating translations that are more aligned with human-written poetic translations and provide an accurate representation of the poem’s meaning.

Recent advancements in the modeling capabilities of large scale models contributed to significant changes in text-to-image synthesis (TIS) \cite{emnlp_peom}. 
However, the quality of images generated by these models heavily relies on properly crafted, keyword-based text prompts \cite{zhong,ecai_poem}. The dependence arises from training data limitations, which require detailed prompts to produce high-quality visuals \cite{dalle}. Building on this idea, we have integrated semantic graph knowledge for the prompt formulation, which serves as input to the image generation framework. By mapping tokens, dependencies, and metaphorical relationships, semantic graphs facilitate the creation of well-structured prompts that generate contextually rich and meaningful images, ensuring that the generated visuals capture the main idea of the poem rather than just its literal words. Figure \ref{motivation_image} illustrates the representation of poems generated by our \textit{TAI} two-step framework. In summary, the key contributions of this study are as follows:
\begin{enumerate}
    \item An Indian poem translation module incorporating the Odds Ratio Preference Optimization algorithm to produce translations that closely align with human-written translations.
    \item A semantic graph-based approach for image prompt construction that captures the themes, metaphorical meanings, and visual elements present in poetry.
    \item A well-curated \textit{Morphologically Rich Indian Language Poems (MorphoVerse)} dataset with 1,570 poems in 21 diverse Indian languages.
    \item Culturally informed human evaluations, where experts from diverse linguistic backgrounds confirmed that our framework generates images that best represent poems in terms of meaning, cultural themes, and visual elements.
\end{enumerate}

\section{Background and Related Works}


\subsubsection{MultiLingual Poetry Translation: } Early research in poetry machine translation focused on phrase-based translation systems, with initial efforts aimed at translating French poetry into metrical English verse \cite{ref10}. Later, \cite{ref11} applied statistical methods to translate rhymed poetry, successfully converting Italian poetry into English. \cite{ref12} explored automatic sense disambiguation for translating Hindi poetry into Dogri, leveraging the n-gram approach for accurate word mapping. \cite{ref13} developed a rule-based machine translation system that used declension rules to convert English sentences into Hindi, with grammatical structure verification through the Stanford POS tagger. More recently, \cite{ref14} introduced a Hybrid Machine Translation (HBMT) model for converting Hindi poetry into English, significantly improving semantic and syntactic accuracy in poetry translation. 
While previous research has made major improvements in multilingual poetry translation, limited work has been done on low-resource poetry in various Indian languages. To narrow this gap, we focus on translating low-resource Indian poetry into English from a curated poetry dataset that includes poems from 21 different Indian languages.


\subsubsection{Image Generation via Diffusion Models: } Recent advancements in diffusion models \cite{ref15,ref16,ref17} have significantly enhanced the quality of text-to-image generation, with models such as Dreambooth \cite{dreambooth} and DALLE 3 \cite{dalle} achieving impressive results. Additionally, several recent studies  \cite{ref21,ref22,ref23,jamil2025poetry,ecai_poem} have explored image understanding feedback, leveraging reward-based models to refine diffusion techniques for better text-image alignment. Despite these advancements, current models still struggle with complex prompts, often generating images that lack core elements, leading to semantic confusion \cite{limi1,lim2,lim3}. To address these issues, we incorporated a semantic graph for prompt construction to capture the semantic meanings of the poem text.

\section{{\em MorphoVerse} Dataset}
In the existing literature, datasets available for poetry generation in Indian languages such as Hindi Poems \cite{hindipeoms}, Sukhan \cite{sukhan} and Devangri Poem dataset \cite{devanagri} are predominantly in Hindi, with no author translations. To address this gap and achieve our research objectives, we curated the \textit{Morphologically Rich Indian Language Poems (MorphoVerse)} Dataset, comprising 1,570 poems in 21 diverse Indian languages, and provided comparative statistics in Table \ref{dataset}.  We outline the steps below taken to construct our dataset.
\newline\textbf{1. Data Collection :} We compiled the \textit{MorphoVerse} dataset from various open online sources, focusing specifically on Indian languages. To ensure the reliability and accuracy of the data collection process, we selected a group of three English-proficient final-year undergraduate students. These individuals were chosen based on their technical expertise and ability to verify the authenticity of the poems sourced from different online platforms. For low-resource languages, we conducted an extensive search for corresponding author translations across multiple online sources to enhance the dataset’s quality and completeness. 
\newline\textbf{2. Data Cleaning :} Given the diverse sources of the collected poems, we carried out extensive data cleaning to maintain consistency. This process involved, removing duplicate entries, correcting inaccuracies, eliminating extraneous whitespace and HTML tags, and addressing other formatting errors.
After cleaning the data, we carefully reviewed the dataset, and resolved any remaining discrepancies using Cohen's Kappa score to ensure consistency and reliability.

\begin{table}[!ht]
\centering
\setlength{\tabcolsep}{1mm}
\begin{tabular}{@{}lccc@{}}
\toprule
\textbf{Dataset} & Count & Languages & Translations \\ \midrule
\textit{Hindi\_Poems} & 2500 & Hindi & \XSolidBrush \\
\textit{SUKHAN} & 845 & Hindi & \XSolidBrush \\
\textit{Devanagri Poem dataset} & 1500 & Devanagri & \XSolidBrush \\
\textit{MorphoVerse (OURS)} & 1570 & 21 & \Checkmark \\ \bottomrule
\end{tabular}
\caption{Statistics of our Proposed Dataset \textit{MorphoVerse} in Comparison to other Hindi Poems Dataset}
\label{dataset}
\end{table}

\section{Methodology}

\subsection{Problem Statement}

We formalize the poem-to-image generation task as a text to image synthesis task, where a low resource Indian poem \( P \) is transformed into a generated image \( I \). This process consists of three components: (1) A translation module that converts \( P \) into an English language Poem while preserving the morphological essence of the poems (2) A semantic graph, which captures key tokens, dependencies, and metaphorical relationships between poem texts and (3) Creating an appropriate image prompt for generating an image by incorporating both linguistic and semantic knowledge.
Our overall framework is illustrated in Figure \ref{framework}. This section provides a detailed explanation of the Translation Module, the Semantic Graph Construction utilized for Image prompt construction, and the Image Prompt Creation for image generation.

\subsection{Translation Module}
\label{translationmodule}

\begin{figure*}[!ht]
\centerline{\includegraphics[width=\textwidth]{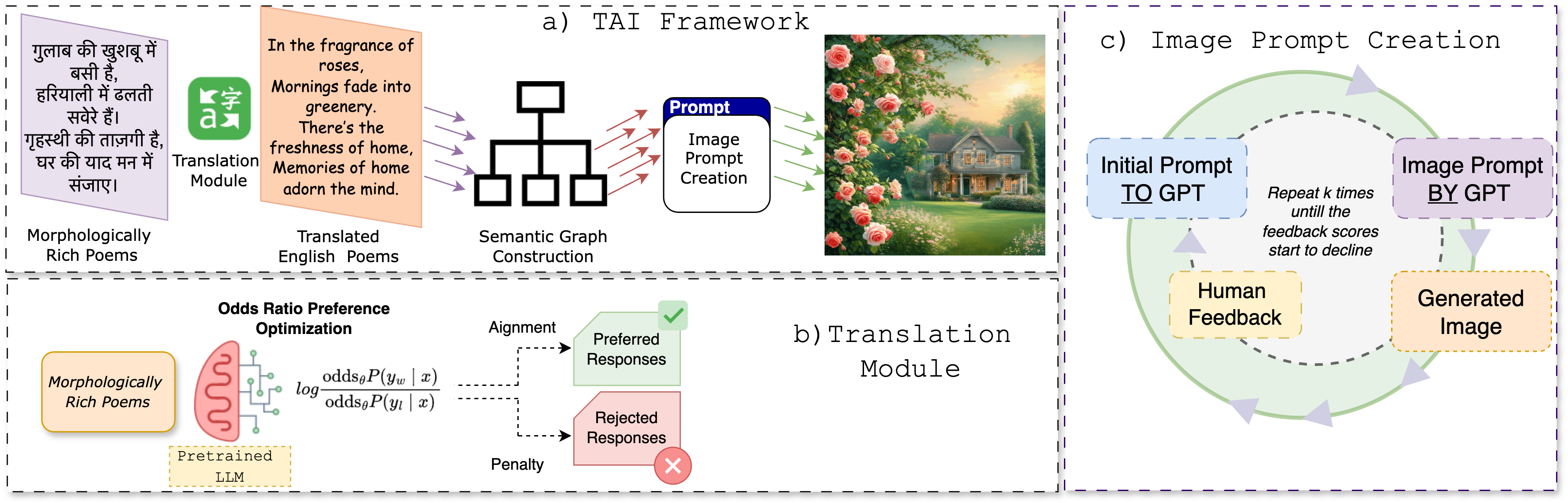}}
\caption{Framework for Translation and Image Generation \textit{TAI} Framework}
\label{framework}
\end{figure*}

Poetry translations must preserve the original text’s poetic essence, rhythm, and stylistic nuances \cite{form_ndform}. To achieve this, we employ Large Language Models (LLMs). However, prior research has shown that widely used models like GPT-3.5 and GPT-4 often produce literal translations, failing to capture the poetic structure and metaphorical depth of the source text \cite{wang2024bestwaychatgpttranslate}. To improve the translation process, we implement Odds Ratio Preference Optimization (ORPO) \cite{orpo} to refine the translation process and ensure that the model prioritizes poetically meaningful outputs over literal translations. As illustrated in Figure \ref{framework} (b), the alignment algorithm introduces an odds ratio-based penalty to the conventional negative log-likelihood loss, enabling the model to distinguish between preferred and unpreferred translation styles. The odds of generating a translated poem \( y \) given an input poem \( x \) under model \( \theta \) is defined as:

\begin{equation}
\text{odds}_\theta(y \mid x) = \frac{P_\theta(y \mid x)}{1 - P_\theta(y \mid x)}
\end{equation}

If \(
\text{odds}_\theta(y \mid x) = k
\), it means that the model \( \theta \) is k times more likely to generate the output poem \( y \) than to not generate it. To reinforce alignment, we introduce the odds ratio (OR) to compare the preferred response (\( y_w \)) with the non-preferred response (\( y_l \)). 


\begin{equation}
\text{OR}_\theta(y_w, y_l) = \frac{\text{odds}_\theta(y_w \mid x)}{\text{odds}_\theta(y_l \mid x)}
\end{equation}
The ORPO objective function combines the standard Supervised Fine-Tuning (SFT) loss with an additional term that penalizes the model if it fails to differentiate sufficiently between favored and disfavored responses as expressed in Equation \ref{orpoobjective}. The hyperparameter \( \lambda \) regulates the intensity of this penalty. 
\begin{equation}
L_{\text{ORPO}} = \mathbb{E}_{(x, y_w, y_l)} \left[ L_{\text{SFT}} + \lambda \cdot L_{\text{OR}} \right].
\label{orpoobjective}
\end{equation}

The penalty term \( L_{\text{OR}} \) is based on the log odds ratio, refined by a sigmoid function to smooth the gradient.

\begin{equation}
L_{\text{OR}} = - \log \sigma \left( \log \frac{\text{odds}_\theta(y_w \mid x)}{\text{odds}_\theta(y_l \mid x)} \right)
\end{equation}


\subsection{Semantic Graph Generation}
\label{semanticgraph}

Poetry frequently conveys abstract and hidden meanings beyond the literal meanings of the words. It is essential to capture these subtle details to ensure an accurate visual representation. To approach this issue, we introduce a module designed for the generation of semantic graphs generated from translated poems, thus simplifying the extraction of conceptual relationships and syntactic structures. Given a translated poem \( T \), we define a directed semantic graph \( G = (V, E) \), where \( V \) represents the set of nodes and \( E \) represents the set of edges. Each node is defined as a pair, \(
v_i = (\ell_i, s_i)
\), where \( \ell_i \) denotes the lemma (base form) of token \( t_i \) and \( s_i \) represents its wordnet identifier to resolve contextual meaning (synset). A node is added to the graph for each token:

\begin{equation}
V = \{ v_i \mid v_i = (\ell_i, s_i), t_i \in T \}
\end{equation}

Edges in the graph encode syntactic dependencies and hypernym relations. For dependency relations, an edge is formed between two nodes \( v_i \) and \( v_j \) if token \( t_i \) depends on token \( t_j \), \(
E_{\text{dep}} = \{ (v_i, v_j) \mid t_i \text{ depends on } t_j \}
\) and hypernym edges link a node \( v_i \) to another node \( v_k \) if the synset of \( v_i \) has a hypernym \( s_k \), \(
E_{\text{hypernym}} = \{ (v_i, v_k) \mid s_i \in \text{Hypernyms}(s_k) \}
\). In other words, \(E_{\text{dep}}\) captures direct grammatical relationships between words, while
\(E_{\text{hypernym}}\) connects words to their more general meanings. Our approach further implements a greedy modularity optimization \cite{al2018greedy} to identify semantically coherent clusters within the graph. Let \( G' = (V', E') \) be the undirected version of the graph \( G \). Community detection is formulated as partitioning \( V' \) into clusters \( C = \{ C_1, C_2, \dots, C_m \} \), optimizing the modularity \( Q \):

\begin{equation}
Q = \frac{1}{2m} \sum_{i,j} \left( A_{ij} - \frac{k_i k_j}{2m} \right) \delta(c_i, c_j)
\end{equation}

where \( A_{ij} \) is the adjacency matrix of \( G' \), \( k_i \) and \( k_j \) are the degrees of nodes \( i \) and \( j \), \( m \) represents the total number of edges, and \( \delta(c_i, c_j) \) is 1 if nodes \( i \) and \( j \) belong to the same community and 0 otherwise. 


\subsection{Image Prompt Construction}
\label{promptselection}

The semantic graph constructed from poems is utilized as input for an instruction generation module, which guides the image creation process. This module employs GPT-4o mini for formulating text-based prompts and the Stable-Diffusion-3.5-Medium \cite{stable__diff} for efficient, cost-effective image generation. As illustrated in Figure \ref{framework} 
(c), the initial prompt given to GPT is: 
\textit{``Identify the central theme or metaphor from the graph representing a semantic analysis of a poem's text, divided into smaller "communities" or clusters of related words, with their dependencies and relationships illustrated through nodes and edges.
Integrate all key elements into a cohesive visual scene that represents the poem's essence. 
Ensure the generated prompt is coherent, detailed, and suitable for understanding of stable diffusion model. ''}

This initial prompt guides the GPT in creating a detailed image instruction based on the semantic content of the poem. For example, the Image Instruction Prompt generated for the poem shown in Figure \ref{framework}(a) is, \textit{``A serene garden scene at dawn, featuring blooming roses surrounded by lush greenery. The air carries a fragrant bouquet, evoking memories of home. A gentle morning light casts a fresh glow over the landscape, symbolizing the essence of nostalgia and warmth. Include elements of freshness and calm, illustrating the beauty and simplicity of home in nature.''} This prompt is then fed into the image generation model to produce the image. To evaluate the quality and authenticity of the generated images, we incorporated human feedback. We collaborated with four domain experts from the Indian Poetry Society to evaluate the generated images. We sampled 5\% of poems from the \textit{MorphoVerse} dataset, with experts rating each image on a scale of 1 to 5. The average score was calculated after each iteration, with higher scores indicating better alignment with the poem, and lower scores reflecting misalignment. Based on the expert feedback, as shown in Figure \ref{framework}(c), we iteratively refined the GPT prompts, improving the generated image instructions and producing more accurate and aligned images. This iterative refinement process resulted in a gradual improvement in the generated images and their alignment with poems. After five rounds of feedback, the scores began to decline in the sixth round. At this point, we finalized the fifth prompt as \textit{``From the graph representing a poem’s semantic analysis, extract key metaphors and thematic elements. Develop a detailed image generation prompt that incorporates all significant nodes and their relationships into a unified visual scene. The instruction should be imaginative, coherent, and tailored for Stable Diffusion’s interpretation of metaphorical imagery.''} Based on this
prompt, we use GPT to generate detailed image instruction \( I \). The final image is generated by a Stable Diffusion Model, which takes the generated instruction \( I \) as input. Thus, for each poem in the dataset \( \mathcal{D} = \{P_1, P_2, ..., P_n\} \), we obtain a corresponding set of generated images \( \mathcal{I} = \{I_1, I_2, ..., I_n\} \). Based on this
prompt, we use GPT to generate detailed image instructions \( I \) , where \( I \) is a structured prompt containing descriptive elements extracted from semantic clusters in \( G \). The final image is generated by a Stable Diffusion Model, which takes the generated instruction \( I \) as input. Thus, for each poem in the dataset \( \mathcal{D} = \{P_1, P_2, ..., P_n\} \), we obtain a corresponding set of generated images \( \mathcal{I} = \{I_1, I_2, ..., I_n\} \). 


\section{Experiments}

\textbf{Training Settings} We fine-tuned the baseline models on 1,570 samples from the \textit{MorphoVerse}  Dataset, with a 70:30 split between training and validation. Fine-tuning was performed using LoRA \cite{LoRa} models with a learning rate of 1e-04 for three epochs, a weight decay of 1e-2, a batch size of 32, and 88,000 training steps. The LoRA rank was set at 32, with alpha = 32. Additionally, Direct Preference Optimization (DPO) \cite{dpo} was applied on the supervised fine-tuned model with a learning rate of 1e-04, and ORPO \cite{orpo} was used with a learning rate of 5e-05 for four epochs on pre-trained models. All experiments were conducted using PyTorch on a single server with NVIDIA RTX 100 GPUs.
\newline\textbf{Baselines} For the translation task, we selected baseline models trained with Indian languages, including Mistral-7B-Instruct-v0.3 \cite{mistral}, Qwen 2.5 \cite{qwen}, gemma-2-9b-it \cite{gemma_2024}, Llama-3.1-8B-Instruct \cite{llama_3}, and sarvam-1 \footnote{\url{https://huggingface.co/sarvamai/sarvam-1}}. For the text-to-image generation task, we employed pretrained diffusion models, including Stable-Diffusion-3.5-Medium \cite{stable__diff}, Playground-V2.5-1024px-Aesthetic \cite{li2024playground}, and Sana\_1600M\_1024px \cite{xie2024sana}.
\newline\textbf{Evaluation Protocols}  To evaluate translation quality, we used ROUGE \cite{lin2004rouge}, BLEU \cite{bleu}, METEOR \cite{meteor}, and COMET \cite{comet} scores. To evaluate alignment between the generated image, poem, and prompt, we used BLIP \cite{blip} Score, where captions were generated for images and their similarity to the original image-generation prompt was computed. Additionally, Long-CLIP \cite{longclip} was used to measure the cosine similarity between the poem and the generated image. We also utilize the Image Rewards metric \cite{imagereward} to further assess the quality of the generated images. This metric evaluates human preferences in text-to-image synthesis through extensive analysis and  providing a quantitative score for the generated images.

\begin{figure}[!htpb]
\centerline{\includegraphics[width=\columnwidth]{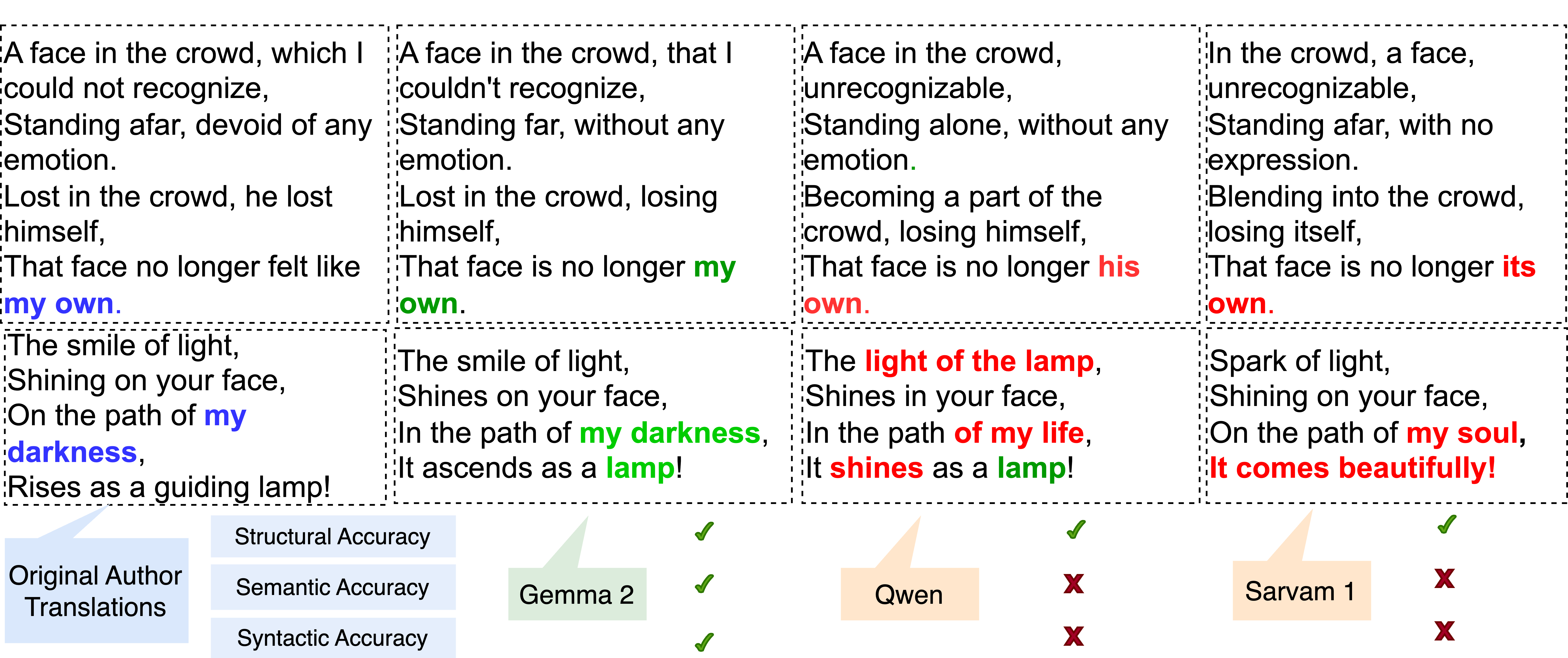}}
\caption{Comparative Qualitative Analysis of Translated Poems Using the odds ratio preference optimization (ORPO) Approach Across
Different LLMs.}
\label{translation_qualitative}
\end{figure}

\begin{figure}[!htpb]
\centerline{\includegraphics[width=\columnwidth]{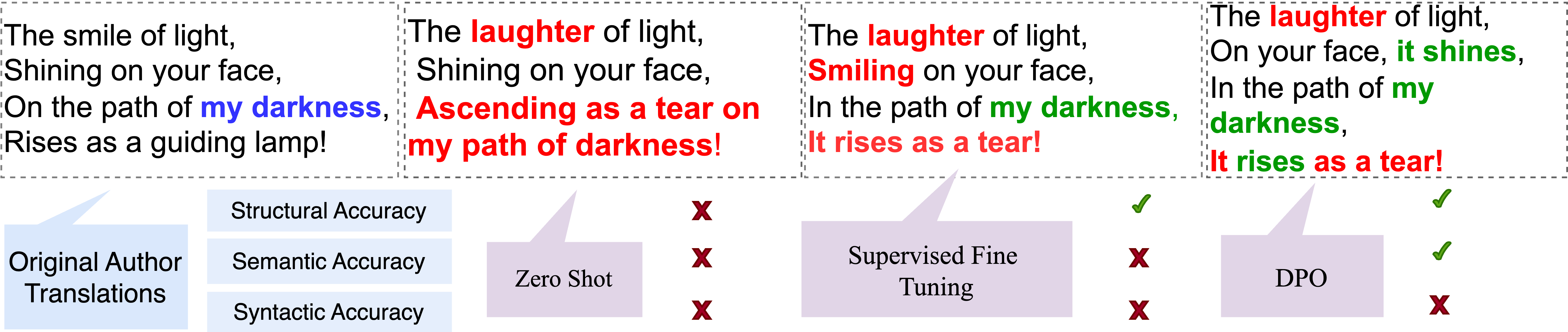}}
\caption{Comparative Qualitative Evaluation of Translated Poems Using Different Approaches with the \textit{Gemma} Model.}
\label{all_gemma}
\end{figure}

\begin{figure*}[!ht]
\centerline{\includegraphics[width=\textwidth]{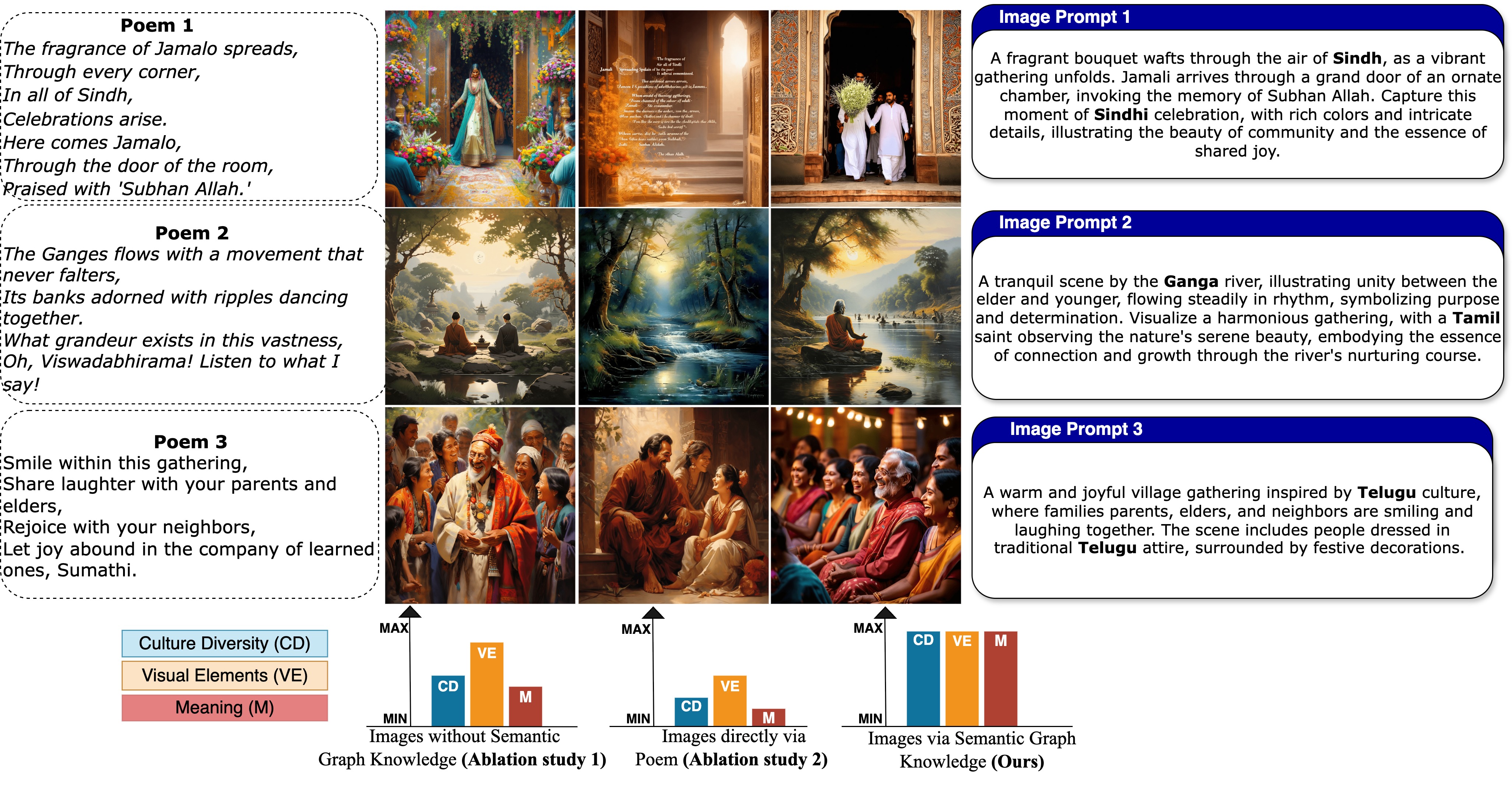}}
\caption{Qualitative Evaluation of the Images Generated using constructed prompts with Semantic graph knowledge}
\label{qualitative}
\end{figure*}
\section{Results}


\subsection{Qualitative Evaluation}
\textbf{Step 1: Translation Task}

We conducted qualitative comparisons of the translated poems, evaluating them based on three key criteria: structural accuracy (whether the number of lines remained consistent), semantic accuracy (whether the meaning was preserved or altered), and syntactic accuracy (whether words were replaced with synonyms or altered in a way that changed the original meaning). As shown in Figure \ref{translation_qualitative}, we established a comparative analysis across multiple models, including Gemma, Mistral, Qwen, and LLama and sarvam, all integrated with the Odds Ratio Preference Optimization (ORPO) approach. Among these, the Gemma model with ORPO demonstrated the highest structural, semantic, and syntactic accuracy when compared to the original English translations. On the other hand, for performing qualitative analysis of the translated poems without the ORPO approach, we show the results of the Gemma model in Figure \ref{all_gemma}. It is observed that the poem generated by the Gemma pretrained model failed to retain structural, semantic, or syntactic accuracy in the pretrained version. After applying the fine-tuning approach, structural accuracy improved, but semantic and syntactic accuracy showed no improvements. On the other hand, the Direct Preference Optimization technique improved both structural and semantic accuracy but failed to improve syntactic accuracy. All the results are highlighted in the table for clearer understanding.

 \textbf{Step 2: Image generation task} We conducted a qualitative analysis of the generated images based on three key aspects: 
\textbf{a) Meaning Capture: } Since poetry is a complex art form, it is crucial that the generated images effectively convey the meaning embedded in the text. As observed in Figure \ref{qualitative}, \textit{Poem 3} evokes nostalgia, serenity, and a deep connection to home, reminiscing about peaceful mornings that bring comfort. The images generated using semantic graph knowledge successfully depicts a home at dawn, surrounded by a rose garden, precisely aligning with the poet’s intent of portraying home as a cherished memory. Similarly, in \textit{Poem 2}, which conveys themes of hope and companionship, the image generated through semantic graph knowledge accurately represents the idea of moving forward to face the world with hope and companionship. A similar portrayal of celebrations is observed in \textit{Poem 1}, demonstrating that the poetic meaning is effectively captured.
\textbf{b) Visual Element Representation:} As shown in Figure \ref{qualitative}, images generated using semantic knowledge successfully incorporate key visual elements, such as trees, the moon, and human figures representing companionship in \textit{Poem 2}. For \textit{Poem 3}, the images accurately depict a home and a rose garden in the morning. All the visual elements mentioned in the poem, essential for representing the image, are effectively captured. Similarly, for \textit{Poem 1}, we observe that doors and celebratory props are accurately portrayed.
\textbf{c) Cultural Representation:}  Among all approaches, semantic graph knowledge demonstrated the strongest ability to capture cultural elements, as seen in \textit{Poem 1} of Figure \ref{qualitative}. This poem, written in Urdu, celebrates the culture of \textit{Sindh}. The generated image accurately portrays this by depicting a man dressed in traditional Sindhi attire engaged in a celebratory moment. This level of cultural representation was absent in images produced through other approaches. 

\subsection{Automated Evaluation}

\textbf{Step 1: Poem Translation}
For the automatic evaluation of poem translation tasks, we tested four different approaches across all LLMs: a) zero-shot translation, b) supervised fine-tuning, and alignment techniques such as c) Direct Preference Optimization (DPO) and d) Odd Ratio Preference Optimization (ORPO). As shown in Table \ref{Translation}, Gemma 2 consistently outperforms all other models across all evaluation metrics, both in zero-shot and optimized settings. Moreover, as observed from Table \ref{Translation}, fine-tuning the model significantly enhances translation quality for each model when compared to the zero-shot setting, demonstrating the model’s ability to effectively learn from additional training data. Furthermore, ORPO consistently improves translation performance across all models, resulting in significant increases in BLEU, METEOR, and COMET scores. Additionaly, we also evaluated the SARVAM model, which is specifically optimized for Indian languages. However, it underperforms compared to other models in both zero-shot and fine-tuned settings, indicating limitations in its translation capabilities for poetry.
\begin{table*}[!ht]
\centering
\begin{tabular}{@{}clccccccccc@{}}
\toprule
Models & \multicolumn{1}{c}{} & \textit{R1} & \textit{R2} & \textit{RL} & \textit{BLEU1} & \textit{BLEU2} & \textit{BLEU3} & \textit{BLEU4} & \textit{METEOR} & \textit{COMET} \\ \midrule
\multirow{4}{*}{\textbf{Qwen}} & \textit{Zero Shot} & 0.453 & 0.2225 & 0.397 & 0.3062 & 0.1988 & 0.1311 & 0.0828 & 0.3639 & -0.1964 \\
 & \textit{Fine Tuned} & 0.5951 & 0.3169 & 0.5264 & 0.4509 & 0.3196 & 0.234 & 0.1674 & 0.4561 & 0.2223 \\
 & \textit{DPO} & 0.5975 & 0.3253 & 0.5302 & 0.4564 & 0.3287 & 0.2449 & 0.1776 & 0.4621 & 0.2327 \\
 & \textit{ORPO} & 0.6163 & 0.3584 & 0.5548 & 0.4831 & 0.3626 & 0.2767 & 0.2099 & 0.4866 & 0.2564 \\ \midrule
\multirow{4}{*}{\textbf{Llama 3.1}} & \textit{Zero Shot} & 0.496 & 0.242 & 0.4285 & 0.3429 & 0.2257 & 0.1519 & 0.0971 & 0.3601 & -0.0842 \\
 & \textit{Fine Tuned} & 0.6356 & 0.3735 & 0.5712 & 0.4918 & 0.3681 & 0.2855 & 0.2189 & 0.4948 & 0.3141 \\
 & \textit{DPO} & 0.6363 & 0.3803 & 0.5731 & 0.4421 & 0.3241 & 0.2441 & 0.1784 & 0.4535 & 0.2946 \\
 & \textit{ORPO} & 0.6421 & 0.3834 & 0.5791 & 0.5038 & 0.3802 & 0.296 & 0.2277 & 0.5094 & 0.3168 \\ \midrule
\multirow{4}{*}{\textbf{Gemma 2}} & \textit{Zero Shot} & 0.6701 & 0.4106 & 0.6101 & 0.5193 & 0.3976 & 0.3109 & 0.2389 & 0.5344 & 0.3579 \\
 & \textit{Fine Tuned} & 0.676 & 0.4229 & 0.6165 & 0.5353 & 0.4148 & 0.3289 & 0.2576 & 0.5452 & 0.3687 \\
 & \textit{DPO} & 0.6677 & 0.4106 & 0.606 & 0.5227 & 0.4018 & 0.3161 & 0.2458 & 0.5343 & 0.3515 \\
 & \textit{ORPO} & \textbf{0.6922} & \textbf{0.4451} & \textbf{0.634} & \textbf{0.5589} & \textbf{0.4421} & \textbf{0.3586} & \textbf{0.2864} & \textbf{0.5693} & \textbf{0.4034} \\ \midrule
\multirow{4}{*}{\textbf{Mistral}} & \textit{Zero Shot} & 0.5056 & 0.238 & 0.4438 & 0.3664 & 0.2335 & 0.1518 & 0.1004 & 0.3621 & -0.0886 \\
 & \textit{Fine Tuned} & 0.6076 & 0.3467 & 0.5445 & 0.4776 & 0.353 & 0.2683 & 0.2036 & 0.4807 & 0.2134 \\
 & \textit{DPO} & 0.6056 & 0.3443 & 0.5434 & 0.4769 & 0.3526 & 0.2697 & 0.2031 & 0.4801 & 0.2264 \\
 & \textit{ORPO} & 0.58 & 0.3249 & 0.5251 & 0.4545 & 0.3318 & 0.2493 & 0.1858 & 0.4521 & 0.131 \\ \midrule
\multirow{4}{*}{\textbf{Sarvam 1}} & \textit{Zero Shot} & 0.2848 & 0.1111 & 0.2329 & 0.1533 & 0.0869 & 0.0427 & 0.017 & 0.1707 & -0.7911 \\
 & \textit{Fine Tuned} & 0.5928 & 0.3242 & 0.5159 & 0.4435 & 0.3276 & 0.2411 & 0.1815 & 0.4514 & 0.2265 \\
 & \textit{DPO} & 0.5934 & 0.3331 & 0.5429 & 0.4672 & 0.3345 & 0.2509 & 0.1819 & 0.4505 & 0.2321 \\
 & \textit{ORPO} & 0.6011 & 0.3348 & 0.5732 & 0.4723 & 0.3546 & 0.2566 & 0.2079 & 0.4625 & 0.2557 \\ \bottomrule
\end{tabular}
\caption{
Results for the translation task on the \textit{MorphoVerse} dataset using our \textit{TAI} Framework's translation module. The results are in terms of ROUGE-1 (R1), ROUGE-2(R2), ROUGE-L(RL), BLEU scores, METEOR, and COMET scores.}
\label{Translation}
\end{table*}
\newline \textbf{Step 2: Image Generation}
Our \textit{TAI} approach consistently outperforms other approaches across all baselines and scores, as demonstrated in Table \ref{quanitative_image}. Given that the Long-CLIP score reflects semantic consistency between text and image, it is evident that sending the constructed prompts with semantic graph knowledge to Stable Diffusion yields the highest score for the constructed prompt (finalized via human feedback in Section Image Prompt Construction). Furthermore, for the Image Rewards metric, we evaluated the alignment between prompts used for image generation and the resulting images using the reward model, where our \textit{TAI} approach demonstrated superior performance. Additionally, we assessed image rewards between the translated poem and the generated image, finding that the generated image closely aligns with the poems. Our \textit{TAI} approach is broadly applicable to all SD-style models. The experimental results from SD 3.5 Medium and Playground v2.5 confirm the flexibility of our designed prompts to other diffusion based Text to Image models.
\begin{table}[!ht]
\centering
\setlength{\tabcolsep}{1mm}
\begin{tabular}{@{}llccc@{}}
\toprule
\multicolumn{1}{c}{Models} &  & \textit{Long-CLIP} & \textit{BLIP} & \textit{IR} \\ \midrule
\multirow{3}{*}{SD 3.5 medium} & \textit{CP} & \textbf{0.2436} & \textbf{0.4613} & \textbf{0.5342} \\
 & \textit{AS 1} & 0.2329 & 0.4510 & 0.3842 \\
 & \textit{AS 2} & 0.2341 & 0.3813 & -0.2471 \\ \midrule
\multirow{3}{*}{Sana 1.6B} & \textit{CP} & \textbf{0.2211} & \textbf{0.4372} & \textbf{0.3638} \\
 & \textit{AS 1} & 0.2222 & 0.4154 & 0.3010 \\
 & \textit{AS 2} & 0.2045 & 0.2648 & -0.3867 \\ \midrule
\multirow{3}{*}{Playground v2.5} & \textit{CP} & \textbf{0.2332} & \textbf{0.4606} & \textbf{0.4684} \\
 & \textit{AS 1} & 0.2201 & 0.4436 & 0.4014 \\
 & \textit{AS 2} & 0.2204 & 0.3198 & -0.2278 \\ \bottomrule
\end{tabular}
\caption{Results for the image generation task for poems in the \textit{MorphoVerse} dataset under different settings : Constructed Prompt(CP), Ablation Study 1(AS 1), Ablation Study 2(AS 2).}
\label{quanitative_image}
\end{table}


\subsection{Ablation Study}
To assess the effectiveness of our proposed approach, we evaluated the poem-to-image generation task under different settings. 
\newline\textbf{Ablation Study 1: Generating Images Using Direct Prompts } In this approach, we provided the translated poem to GPT and prompted it to generate descriptive prompts essential for image generation. The images were generated using the GPT-generated prompts. However, these images failed to accurately capture the meaning and cultural diversity of the poems, as evident from Figure \ref{qualitative} for \textit{Poem 1}. The generated image for \textit{Poem 1} does not effectively capture the \textit{sindh culture}.  
For \textit{Poem 2}, the generated image depicts \textit{a lonely man walking through a forest}, which is different from the poem's intended meaning. Moreover, for \textit{Poem 3}, a house with a rose garden is generated, but it lacks the freshness of the morning, which the poet intended to convey through the lines: \textit{There’s the freshness of home, Memories of home adorn the mind.}
\newline \textbf{Ablation Study 2: Generating Images Directly from the Poem} 
The images generated using this approach did not capture the poems' meaning, visual elements, or cultural diversity. As evident from Figure \ref{qualitative} for \textit{Poem 1}, and \textit{2}, the generated images simply display some poem’s text on a background, lacking any meaningful visual representation. For \textit{Poem 3}, only \textit{rose} is generated, which does not effectively convey the intended message of the poem. Table \ref{quanitative_image} further demonstrates that this approach resulted in a decrease in Long-CLIP and BLIP scores, indicating poor alignment between the generated images and poem verses. Moreover, the image reward score is significantly low, indicating that generating images directly from the poem text produces inadequate and unreliable results. This suggests that text to image generation models lack an inherent understanding of poetic imagery.



\section{Conclusion}

In this paper, we introduce a two-stage \textit{Translation and Image Generation (TAI)} framework designed to generate images for low-resource Morphologically rich Indian language poems. Our approach significantly improves both the translation quality of language models and the accuracy of image generation for poetic texts. To achieve this, we integrate semantic graph knowledge to construct prompts for precise visual representations of poetic verses. Additionally, we implement prompt alignment through human feedback, refining a single prompt given to LLMs to incorporate semantic graph knowledge into image prompt construction. Our method improves image quality while simplifying prompt creation for text-to-image models, reducing user trial and error. It effectively balances semantic meaning, visual elements, and cultural diversity to ensure that the generated images accurately capture the main idea of the poem. Furthermore, we introduce \textit{MorphoVerse}, a dataset comprising 1,570 poems across 21 diverse Indian languages, designed to facilitate research in poetry translation and image generation. Extensive human and quantitative evaluations validate the effectiveness of our \textit{TAI} framework, demonstrating its superiority in creating high-quality, contextually rich images for poetic texts.\footnote{All the experimental research and analysis were conducted in an academic setting at IIT Patna.}

\bibliography{aaai2026}

\end{document}